\documentclass[11pt]{article}

\usepackage[final]{acl}
\usepackage{float}
\usepackage{times}
\usepackage{latexsym}
\usepackage{amsmath}
\usepackage[most]{tcolorbox}
\usepackage{enumitem}
\usepackage[T1]{fontenc}
\usepackage{makecell}
\usepackage[utf8]{inputenc}

\usepackage{microtype}
\usepackage{caption}
\usepackage{inconsolata}
\usepackage{graphicx}
\usepackage{booktabs}
\usepackage{multirow}
\usepackage{colortbl}
\usepackage{xcolor}
\usepackage{enumitem}
\title{Structured Episodic Event Memory}

\author{
  {\bf Zhengxuan Lu}$^{1,3}$, 
  {\bf Dongfang Li}$^2$, 
  {\bf Yukun Shi}$^2$,\\
  {\bf Beilun Wang}$^1$,
  {\bf Longyue Wang}$^4$, 
  {\bf Baotian Hu}$^{2,3}$ \\
  \textsuperscript{1}Southeast University, Nanjing, China\\
  \textsuperscript{2}Harbin Institute of Technology (Shenzhen), Shenzhen, China \\
  \textsuperscript{3}Shenzhen Loop Area Institute, Shenzhen, China \\
  \textsuperscript{4}Alibaba Group, Hangzhou, China\\
  \texttt{230249730@seu.edu.cn, lidongfang@hit.edu.cn}
}

\begin{document}
\maketitle
\begin{abstract}
Current approaches to memory in Large Language Models (LLMs) predominantly rely on static Retrieval-Augmented Generation (RAG), which often results in scattered retrieval and fails to capture the structural dependencies required for complex reasoning. For autonomous agents, these passive and flat architectures lack the cognitive organization necessary to model the dynamic and associative nature of long-term interaction. To address this, we propose \textbf{S}tructured \textbf{E}pisodic \textbf{E}vent \textbf{M}emory (\textbf{SEEM}), a hierarchical framework that synergizes a graph memory layer for relational facts with a dynamic episodic memory layer for narrative progression. Grounded in cognitive frame theory, SEEM transforms interaction streams into structured Episodic Event Frames (EEFs) anchored by precise provenance pointers. Furthermore, we introduce an agentic associative fusion and Reverse Provenance Expansion (RPE) mechanism to reconstruct coherent narrative contexts from fragmented evidence. Experimental results on the LoCoMo and LongMemEval benchmarks demonstrate that SEEM significantly outperforms baselines, enabling agents to maintain superior narrative coherence and logical consistency.
\end{abstract}

\section{Introduction}\label{sec:intro}

Large Language Models (LLMs) have evolved into sophisticated agents capable of complex reasoning and long-term interaction~\citep{achiam2023gpt, xi2025rise}. However, LLM-based agents remain limited by their finite context windows and the lack of a stable long-term memory system~\citep{DBLP:journals/corr/abs-2310-08560}. This constraint causes reasoning capabilities to degrade over extended sessions, as the agent cannot effectively recall critical information once it exceeds the immediate context. Developing a robust long-term memory is therefore a central challenge in building autonomous agents.

\begin{figure*}[ht!]
    \centering
    \includegraphics[width=0.95\textwidth]{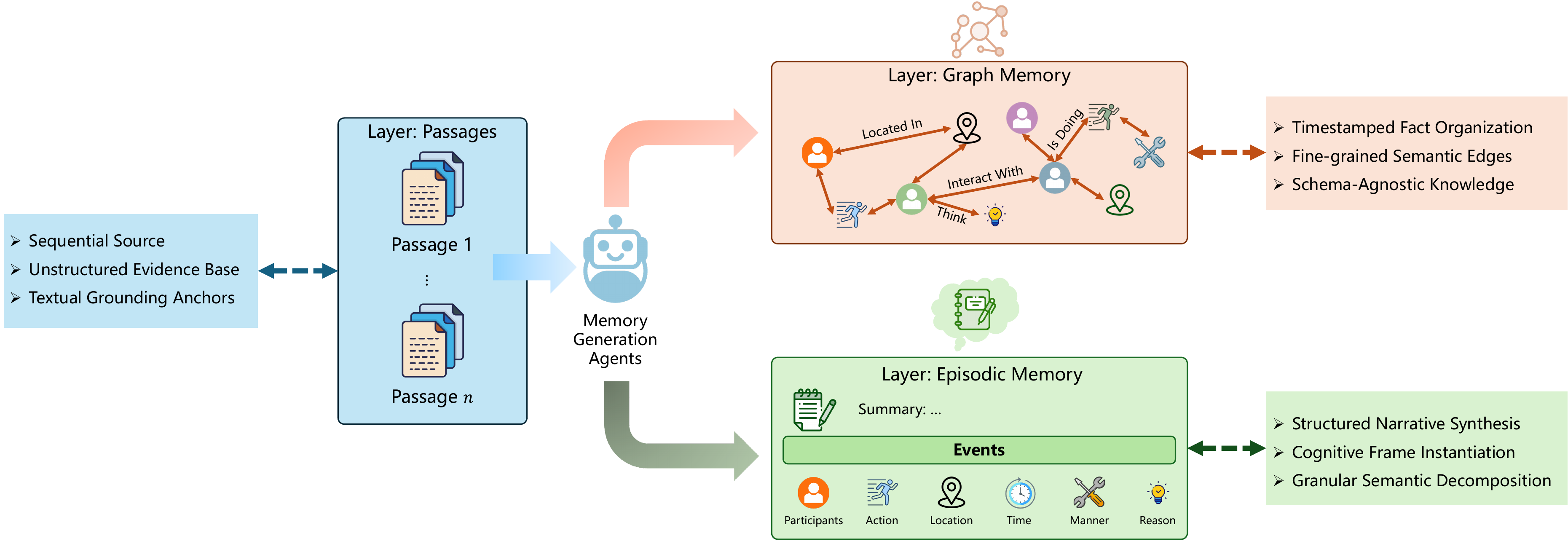}
\caption{\textbf{Overview of the SEEM hierarchical memory architecture.} The system transforms unstructured interaction passages into a dual-layer representation, integrating a semantic Graph Memory Layer for static facts with a structured Episodic Memory Layer for event-centric details. This hierarchical design enables the agent to effectively synergize stable factual knowledge with dynamic narrative contexts for coherent long-term reasoning.}
\vspace{-4mm}
    \label{fig:memory_pipeline}
\end{figure*}

To address this, Retrieval-Augmented Generation (RAG) has emerged as a standard paradigm to supplement LLMs with external knowledge~\citep{lewis2020retrieval}. Traditional RAG systems rely on vector similarity to retrieve local text passages~\citep{karpukhin2020dense}. While efficient, they often struggle with multi-hop reasoning tasks that require understanding the structural dependencies between disparate facts. Recent advancements, such as GraphRAG~\citep{edge2024local} and Mem0~\citep{chhikara2025mem0}, attempt to solve this by organizing information into graph databases. Nevertheless, these approaches face significant structural limitations. Most existing systems rigidly bind semantic content to fixed graph structures or predefined schemas. This rigidity mitigates the memory from dynamically reorganizing as new knowledge arrives. Consequently, these systems frequently suffer from scattered retrieval~\cite{gutierrez2025from}, where the retrieved context is fragmented into isolated pieces, failing to provide the coherent narrative required for complex reasoning.

% To bridge this gap, we propose \textbf{S}tructured \textbf{E}pisodic \textbf{E}vent \textbf{M}emory (\textbf{SEEM}), a hierarchical framework designed for dynamically evolving agentic memory. Unlike static retrieval systems, SEEM introduces a dual-layer architecture that combines a schema-based knowledge graph with a dynamic episodic memory layer. Grounded in cognitive frame theories~\citep{minsky1975framework, fillmore1976frame}, SEEM transforms unstructured interaction streams into discrete \textit{Episodic Event Frames} (EEFs). These frames act as structured cognitive units that organize interactions into queryable dimensions, such as participants, actions, and causality. Furthermore, SEEM employs an \textit{Associative Fusion} mechanism that conceptually aligns with human memory consolidation. This allows the system to synthesize related events into coherent scenes while maintaining the link back to the raw evidence. By doing so, the agent can evolve its understanding of events dynamically and maintain narrative integrity over long horizons. 

To bridge this gap, we propose \textbf{S}tructured \textbf{E}pisodic \textbf{E}vent \textbf{M}emory (\textbf{SEEM}), a hierarchical framework that transforms continuous interaction streams into a cohesive dual-layer architecture. This system is composed of an Episodic Memory Layer (EML), which captures dynamic narrative progression by extracting and fusing structured Episodic Event Frames (EEFs) inspired by cognitive frame theories~\citep{minsky1975framework, fillmore1976frame}, and a complementary Graph Memory Layer (GML) that organizes static factual details into a relational graph. Both layers are anchored to their original source passages via precise provenance pointers, which ensures that abstract memory units remain traceable to raw passages. During inference, these layers are synergized through a hybrid retrieval process utilizing a Reverse Provenance Expansion (RPE) mechanism, allowing the agent to reconstruct a coherent and logically consistent context for complex reasoning.
Extensive experiments are conducted on the LoCoMo~\citep{maharana2024evaluating} and LongMemEval~\citep{wu2025longmemeval} benchmarks. Our results demonstrate that SEEM consistently outperforms competitive memory-augmented and dense retrieval baselines. Notably, it surpasses HippoRAG 2~\cite{gutierrez2025from} by an absolute margin of 4.4\% on LongMemEval. Moreover, supplemental tests under incremental construction settings confirm its stability and robustness for real-world sequential deployment. 
% We conduct extensive experiments on the LoCoMo~\cite{maharana2024evaluating} and LongMemEval~\cite{wu2025longmemeval} benchmarks. SEEM outperforms competitive memory-augmented and dense retrieval baselines. Notably, it surpasses HippoRAG-2~\cite{gutierrez2025from} by an absolute margin of 4.4\% on LongMemEval and demonstrates exceptional robustness in adversarial reasoning tasks compared to competitive baselines. 

Our contributions are summarized as follows:

\begin{itemize}
    \item We introduce SEEM, a hierarchical framework that synergizes GML for relational facts with EML to capture dynamic narrative progression.
    \item  We propose the EEFs and RPE mechanism, which transform interaction passages into multi-attribute cognitive units linked by provenance pointers to mitigate the scattered retrieval problem.
    \item We provide extensive empirical validation demonstrating that SEEM outperforms competitive memory-augmented and dense retrieval baselines in maintaining logical consistency and narrative coherence.
\end{itemize}
\section{Related Work}

\textbf{Vector-based RAG.} 
RAG addresses the parametric constraints of LLMs by accessing external corpora via vector similarity~\cite{lewis2020retrieval}. However, standard RAG systems predominantly rely on flat vector spaces, which operate in a de-contextualized manner~\cite{gao2023retrieval}. This often fails to capture the structural dependencies required for complex multi-hop reasoning, resulting in scattered retrieval where the retrieved context lacks the coherence necessary for consistent long-term interactions~\cite{tang2024multihop,gutierrez2025from}.

\paragraph{Structured Semantic Memory.} 
To bridge semantic gaps, structure-augmented approaches organize memory into knowledge graphs or hierarchical summaries. GraphRAG~\citep{edge2024local} and RAPTOR~\citep{sarthi2024raptor} utilize summaries to link related text segments, while HippoRAG 2~\citep{gutierrez2025from} leverages graph algorithms to facilitate associative retrieval. Despite these gains, such methods often suffer from lack of structural differentiation, where high-level thematic abstracts and fine-grained facts are entangled~\cite{edge2024local}. Furthermore, heavy reliance on LLM-generated summarization can introduce noise, causing performance on basic factual tasks to deteriorate compared to standard RAG~\cite{cuconasu2024power,wu2025pandora}.

\paragraph{Episodic Memory.} 
A fundamental distinction exists between general semantic memory and episodic memory grounded in specific spatiotemporal contexts~\cite{tulving1972episodic}. While recent systems such as Mem0~\citep{chhikara2025mem0} and Graphiti~\citep{rasmussen2025zeptemporalknowledgegraph} track interaction histories, they may struggle to preserve coherent event contexts due to selective summarization or rigid entity-centric relations. Specifically, these methods frequently fail to integrate essential situational dimensions, including time, causality, and participants, into a unified representation. Consequently, there remains a need for a hierarchical memory to handle the spatiotemporal dynamics of continuous interactions. In contrast, our proposed framework is designed to address this specific gap.
\section{Methodology}\label{sec:methodology}
The SEEM framework transforms a continuous stream of interaction passages into a hierarchical memory architecture composed of two complementary layers. The Episodic Memory Layer (EML) focuses on capturing the narrative progression by extracting and fusing structured Episodic Event Frames (EEFs) while the Graph Memory Layer (GML) organizes static factual relations into a structured relational graph. Both layers are grounded in original passages through a system of provenance pointers, which maintain the link between abstract memory units and their raw passage. During inference, these layers are integrated through a hybrid retrieval process utilizing the Reverse Provenance Expansion (RPE) mechanism to reconstruct a coherent and logically consistent context.
\subsection{Problem Formulation}\label{subsec:problemformulation}

The task of memory-augmented generation in long-term interactions is defined as follows. Given a chronological sequence of interaction passages $\mathcal{P} = \{p_1, p_2, \dots, p_T\}$, where each passage $p_t$ represents a discrete unit of historical context, and a current user query $q \in \mathcal{Q}$, the objective is to generate a response $a$ that is factually consistent with $\mathcal{P}$ and contextually relevant to $q$.

We formulate this problem as the optimization of a conditional probability $P(a \mid q, \mathcal{P})$. Due to the significant length and semantic density of $\mathcal{P}$, the task requires the construction of an intermediate memory representation $\mathcal{M}$ to bridge the gap between historical evidence and current reasoning. The process is decomposed into two core stages:

\paragraph{Memory Consolidation.} 
We define a transformation function $\Phi: \mathcal{P} \rightarrow \mathcal{M}$ that maps the raw interaction sequence into a structured representation space $\mathcal{M}$. This stage is designed to preserve essential thematic and relational information while mitigating the noise inherent in raw text.

\paragraph{Conditioned Generation.} 
A retrieval augmented generation function $G(q, \mathcal{M}) \rightarrow a$ is employed to identify a relevant subset $\mathcal{M}_{sub} \subseteq \mathcal{M}$ based on the query $q$, leading to the final response generation:
\begin{equation}
    a = \arg\max_{a'} P(a' \mid q, \mathcal{M}_{sub}; \theta)
\end{equation}
where $\theta$ denotes the parameters of the underlying generative model.
Here, the core challenge lies in designing a representation space $\mathcal{M}$ that can effectively encode the narrative continuity and factual dependencies within $\mathcal{P}$. The system must ensure that the transition from $\mathcal{P}$ to $\mathcal{M}$ maintains provenance, allowing the final generation process to be grounded in the original source evidence.

\subsection{Episodic Memory Generation and Fusion}
\label{subsec:episodic_gen_fusion}

To maintain a coherent understanding of long-term interactions, we introduce a structured episodic memory layer. Instead of storing raw interaction turns, we transform a sequence of passages $\mathcal{P} = \{p_1, p_2, \dots, p_T\}$ into discrete, event-centric units. As illustrated in Figure~\ref{fig:memory_pipeline}, this process consists of two phases: (1) extracting structured episodic event frames from each passage and (2) performing associative consolidation to merge related frames.

\begin{figure*}[ht!]
    \centering
    \includegraphics[width=0.9\textwidth]{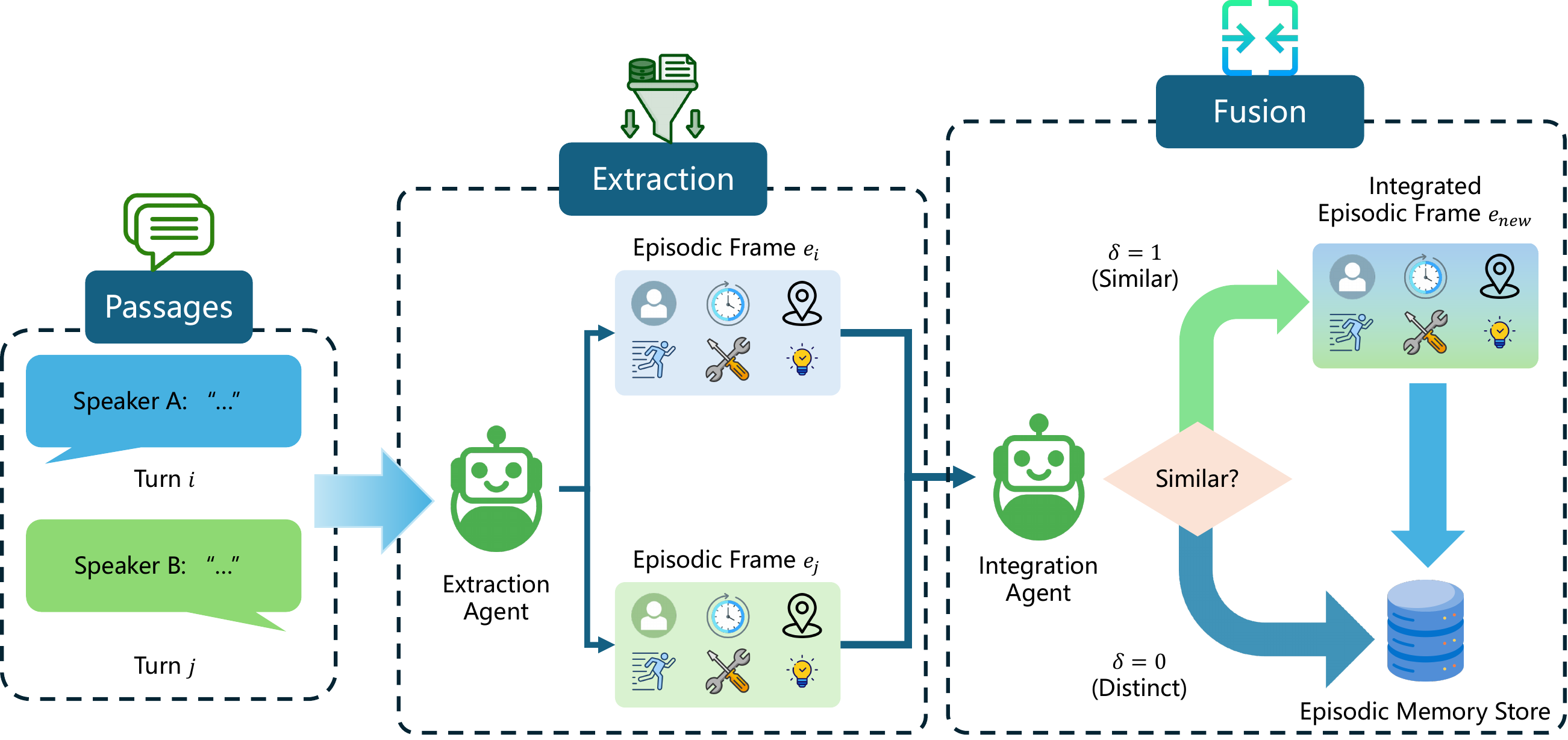}
    \caption{Overview of the associative consolidation and fusion. The $\mathcal{F}_{\text{ext}}$ first transforms raw interaction passages into structured EEFs, which are then processed by $\mathcal{F}_{\text{judge}}$ for the dynamic fusion of semantically related events. This mechanism aligns with associative consolidation to maintain a coherent and synthesized episodic memory store.}
    \vspace{-4mm}
    \label{fig:memory_fusion}
\end{figure*}

\subsubsection{Episodic Event Frame Extraction}
We treat each passage $p_t$ as a source signal to be instantiated into a cognitive frame. Following the principles of frame semantics~\citep{fillmore1976frame}, an EEF $\mathbf{e}_t$ encapsulates the structured semantics of $p_t$. We employ an LLM-based agent, $\mathcal{F}_{\text{ext}}$, to parse $p_t$ into granular semantic roles and a high-level summary. To ensure the abstract memory remains grounded, each frame is linked back to its source passage via a \textit{provenance pointer} $\rho^{eml}_t$. The formal definition is:
\begin{equation}
    \begin{split}
        \mathbf{e}_t &= \mathcal{F}_{\text{ext}}(p_t; \theta) \\
        &= \Big\langle \rho^{eml}_t, v_{\text{sum}}, \Big\{ \big\langle v_{\text{par}}, v_{\text{act}}, v_{\text{tmp}}, \\
        &\qquad\quad v_{\text{spa}}, v_{\text{cau}}, v_{\text{man}} \big\rangle^{(k)} \Big\}_{k=1}^{N_t} \Big\rangle
    \end{split}
\end{equation}
where $v_{\text{sum}}$ is the event summary, and the subsequent components represent semantic roles: Participants ($v_{\text{par}}$), Action ($v_{\text{act}}$), Time ($v_{\text{tmp}}$), Location ($v_{\text{spa}}$), Causality ($v_{\text{cau}}$), and Manner ($v_{\text{man}}$). This hierarchical structure allows the agent to navigate memory through both thematic abstractions and precise textual anchors.

\subsubsection{Associative Consolidation and Fusion}
To mitigate memory fragmentation, we implement an associative fusion mechanism that merges related observations into coherent scenes. When generating a new candidate frame $\mathbf{e}_t$, the system retrieves the most relevant historical frame $\mathbf{e}_{\text{prev}}$ and uses an LLM-based judge to determine if they belong to the same event:
\begin{equation}
    \delta_t \leftarrow \mathcal{F}_{\text{judge}}(\mathbf{e}_t, \mathbf{e}_{\text{prev}} \mid prompt_{\text{sim}})
\end{equation}
If $\delta_t=1$, the integration agent $\mathcal{F}_{\text{fuse}}$ performs an associative merge, synthesizing the attributes of both frames and updating the summary $v_{\text{sum}}$. Note that we aggregate their provenance pointers, updating $\rho^{eml}_t$ to point to the union of all involved source passages. This ensures that a single consolidated frame can later serve as an entry point to all relevant evidence scattered across different turns.

\subsection{Graph Memory Construction}
\label{subsec:graph_memory_construction}

While the EML captures the narrative flow, the GML organizes static facts into a consistent relational structure. 

\subsubsection{Fact Extraction and Grounding}
For each passage $p_t$, the system extracts a set of relational quadruples $\mathcal{K}_t$ to form a schema-agnostic knowledge graph:
\begin{equation}
    \mathcal{K}_t = \{ (s, r, o, \tau) \mid s, o \in \mathcal{E}, r \in \mathcal{R}, \tau \in \mathcal{T} \}
\end{equation}
where $s$ and $o$ are entities, $r$ is the relation, and $\tau$ denotes the temporal validity. Each node in the graph is also linked to its source passage $p$ via \textit{provenance pointers} $\rho^{gml}_t$. To maintain graph integrity, we merge nodes that exceed a vector similarity threshold, bridging lexical variations across different passages.

\subsection{Hybrid Retrieval and Context Integration}
\label{subsec:hybrid_retrieval}

During inference, we integrate the structured facts from the GML with the narrative details from the EML through a multi-stage retrieval process.

\subsubsection{Relational Propagation and Passage Retrieval}
The system initiates retrieval by extracting structured quadruples from the query $q$ to ensure structural alignment with the GML. A shared semantic encoder transforms each query-derived quadruple into a dense vector representation. The retrieval engine then computes the semantic similarity between these query vectors and the pre-indexed embeddings of the facts store within the GML using cosine similarity. By ranking these scores across the relational space, the system identifies the most relevant facts to form the initial seed set $\mathcal{K}_{top}$. We then execute a propagation algorithm~\cite{haveliwala2002topic} using $\mathcal{K}_{top}$ as the seed set to compute a distribution over graph nodes. This relational traversal identifies the set of most relevant initial passages $\mathcal{P}_{ret} = \{p_1, p_2, \dots, p_n\}$ through their provenance pointers.

\subsubsection{Reverse Provenance Expansion}
\label{subsubsec:rpe}
Initial retrieval often suffers from context fragmentation because critical details of an event may be scattered across multiple turns that lack direct lexical overlap with the query. To solve this, we use the EML as a semantic bridge. We first retrieve the event frames associated with the initial passages: $\mathcal{E}_{ret} = \bigcup_{p \in \mathcal{P}_{ret}} \Phi(p)$, where $\Phi(p)$ identifies the frames linked to passage $p$.

We then implement the reverse provenance expansion mechanism. By accessing the aggregated provenance pointers $\rho^{eml}(\mathbf{e})$ of each retrieved frame (as formed during the fusion phase in Section \ref{subsec:episodic_gen_fusion}), we expand the evidence set to include all related passages:
\begin{equation}
    \mathcal{P}_{final} = \mathcal{P}_{ret} \cup \bigcup_{\mathbf{e} \in \mathcal{E}_{ret}} \rho^{eml}(\mathbf{e})
\end{equation}
This ensures that if any fragment of an event is activated, all its constituent textual supports are included in the final context, providing a complete narrative for reasoning.

\subsubsection{Context Synthesis}
The final reasoning context $\mathbf{C}$ is synthesized by serializing the expanded passages $\mathcal{P}_{final}$, the structured event frames $\mathcal{E}_{ret}$, and the relational facts $\mathcal{K}_{top}$. This composite context enables the LLM to resolve temporal ambiguities and maintain logical consistency by cross-referencing high-level facts with nuanced episodic evidence.

Finally, the agent generates the predictive response $a$ by conditioned on the query $q$ and the synthesized context $\mathbf{C}$. We model this process as a sequence generation task, where the LLM acts as a decoder $G$ that maximizes the joint probability of the output tokens:
\begin{equation}
    a = G(q, \mathbf{C}) = \arg\max_{a'} \prod_{i=1}^{|a'|} P(y_i \mid y_{<i}, q, \mathbf{C}; \theta)
\end{equation}
where $y_i$ denotes the $i$-th token of the candidate answer $a'$, and $\theta$ represents the parameters of the generator. By prepending the structured memory evidence directly to the input space, the model can perform integrated reasoning across both episodic and relational knowledge, ensuring that the final output is not only grounded in raw evidence but also guided by the high-level semantic structure captured during the memory construction phase.
\section{Experimental Setup}
\begin{table*}[t]
\centering
\small
\setlength{\tabcolsep}{10pt}

\resizebox{0.95\textwidth}{!}{%
\begin{tabular}{lcccc}
\toprule
\multirow{2}{*}{\textbf{Method}} & \multicolumn{3}{c}{\textbf{LoCoMo}} & \textbf{LongMemEval} \\
\cmidrule(lr){2-4} \cmidrule(lr){5-5}
 & BLEU-1 & F1 & $J$ & Acc. \\
\midrule

% --- 第一类：Dense Retrieval ---
\rowcolor{gray!5}
\multicolumn{5}{l}{\textit{Dense Retrieval}} \\
KaLM-Embedding-V2.5~\citep{zhao2025kalmembeddingv2superiortrainingtechniques} & 44.4 & 47.9 & 64.6 & 55.6 \\ 
NV-Embed-v2~\citep{lee2025nvembed} & 53.0 & 57.9 & 74.7 & 58.4 \\
\midrule

% --- 第二类：Memory-based Frameworks ---
\rowcolor{gray!5}
\multicolumn{5}{l}{\textit{Memory-based Frameworks}} \\
Mem0~\citep{chhikara2025mem0} & 34.2 & 43.3 & 54.1 & 56.7 \\
A-MEM~\citep{xu2025mem} & 45.7 & 44.6 & 61.9 & 55.2 \\
HippoRAG 2~\citep{gutierrez2025from} & 53.8 & 58.3 & 76.2 & 60.6 \\

\rowcolor{gray!15} 
\textbf{SEEM (Ours)} & \textbf{56.1} & \textbf{61.1} & \textbf{78.0} & \textbf{65.0} \\

\bottomrule
\end{tabular}%
}
\caption{Performance comparison on LoCoMo and LongMemEval. The best results are highlighted in \textbf{bold}.}
\label{tab:main_results}
\end{table*}
\subsection{Datasets} 
To rigorously evaluate the long-term memory and reasoning capabilities of our framework, we conduct experiments on two representative benchmarks: (1) \textbf{LongMemEval}~\citep{wu2025longmemeval} serves as a comprehensive testbed for memory-augmented chat assistants, designed to simulate dynamic, evolving user-agent interactions. The dataset comprises 500 manually curated questions that assess five core memory competencies. These include \textit{information extraction} (spanning single-session user, assistant, and preference details), \textit{multi-session reasoning} for synthesizing fragmented information, \textit{temporal reasoning} regarding event timelines, and \textit{knowledge updates} to track changing user states. This benchmark is particularly challenging due to its requirement for maintaining factual consistency across extensible chat histories. (2) \textbf{LoCoMo}~\citep{maharana2024evaluating} focuses on the comprehension of extremely long-term, open-domain conversations. Derived from long-form multi-session dialogues that span up to 32 sessions with an average of 16k tokens, this benchmark provides a rigorous assessment of long-range dependency modeling. We utilize its question answering component, which consists of 1,986 samples categorized into five distinct reasoning types: \textit{single-hop} and \textit{multi-hop} reasoning for context retrieval, \textit{temporal} understanding, \textit{open-domain} knowledge integration, and \textit{adversarial reasoning} to test robustness against hallucinations on unanswerable queries.

\begin{table*}[t]
\centering
\small
\setlength{\tabcolsep}{8pt}
\resizebox{0.95\textwidth}{!}{%
\begin{tabular}{lccccc}
\toprule
\multirow{2}{*}{\textbf{Method}} & \textbf{Multi-hop} & \textbf{Temporal} & \textbf{Open-domain} & \textbf{Single-hop} & \textbf{Adversarial} \\
 & (Count: 282) & (Count: 321) & (Count: 96) & (Count: 841) & (Count: 446) \\
\midrule
A-MEM & 29.4 & 39.7 & 15.0 & 37.6 & 78.3 \\
HippoRAG 2 & 31.9 & 53.4 & \textbf{34.7} & 54.2 & 94.2 \\
\rowcolor{gray!15} 
\textbf{SEEM (Ours)} & \textbf{32.3} & \textbf{54.6} & 26.6 & \textbf{58.2} & \textbf{96.9} \\
\bottomrule
\end{tabular}%
}
\caption{Detailed F1 performance breakdown across five question categories on the LoCoMo benchmark. Sample counts for each category are indicated in parentheses. Best results are highlighted in \textbf{bold}.}
\vspace{-4mm}
\label{tab:category_breakdown}
\end{table*}

\begin{table*}[t]
\centering
\small
\setlength{\tabcolsep}{8pt}

\resizebox{0.95\textwidth}{!}{%
\begin{tabular}{lcccccc}
\toprule
\multirow{2}{*}{\textbf{Method}} & \multicolumn{3}{c}{\textbf{LoCoMo}} & \multicolumn{3}{c}{\textbf{LongMemEval}} \\
\cmidrule(lr){2-4} \cmidrule(lr){5-7}
 & F1 & EM & $J$ & F1 & EM & Acc. \\
\midrule
Context (Relevant Only) & 38.60 & 11.40 & 69.28 & 22.60 & 3.20 & 68.60 \\ 
NV-Embed-v2~\citep{lee2025nvembed} & 46.10 & 29.10 & 51.56 & 16.53 & 2.00 & 57.20 \\
\midrule

\rowcolor{gray!5}
\multicolumn{7}{l}{\textit{Memory-based Frameworks}} \\
HippoRAG 2~\citep{gutierrez2025from} & 44.60 & 27.90 & 50.96 & 34.82 & 25.10 & 46.59 \\

\rowcolor{gray!15} 
\textbf{SEEM (Ours)} & \textbf{48.60} & \textbf{30.80} & \textbf{56.11} & \textbf{45.50} & \textbf{29.60} & \textbf{60.80} \\

\bottomrule
\end{tabular}%
}
\caption{Performance comparison on LoCoMo and LongMemEval using the \texttt{Pangu-Embedded-7B} backbone. The best results are highlighted in \textbf{bold}.}
\label{tab:pangu_results}
\end{table*}

\subsection{Metrics}
\label{subsec:metrics}

We evaluate SEEM using a combination of lexical and semantic metrics to capture both surface-level similarity and high-level factual consistency. For LoCoMo, we employ token-level F1~\citep{maharana2024evaluating} and BLEU-1~\citep{10.3115/1073083.1073135} for lexical comparison. To further assess semantic correctness and factual accuracy, we utilize LLM-as-a-Judge ($J$). Specifically, the judge evaluates model responses using the multi-dimensional evaluation prompts introduced in Mem0~\citep{chhikara2025mem0}, with DeepSeek-V3.2~\citep{deepseekai2025deepseekv32pushingfrontieropen} serving as the underlying scoring engine. For LongMemEval, we strictly adhere to the evaluation protocol described in~\citet{wu2025longmemeval}, which utilizes the LLM to perform binary assessments of answer correctness and reports the resulting accuracy. These metrics collectively provide a rigorous basis for measuring performance across diverse long-term interaction scenarios.
\subsection{Baselines}
We compare SEEM against the following approaches: \textbf{KaLM-Embedding-V2.5}~\citep{zhao2025kalmembeddingv2superiortrainingtechniques} employs a compact decoder-only architecture modified with bidirectional attention and mean-pooling, leveraging high-quality data scaling and advanced training techniques to achieve competitive performance as a versatile and efficient embedding model. \textbf{NV-Embed-v2}~\citep{lee2025nvembed} optimizes a decoder-only LLM architecture by incorporating a latent attention layer and bidirectional attention mechanisms to yield high-performance generalist text embeddings for dense retrieval. \textbf{HippoRAG 2}~\citep{gutierrez2025from} adopts a neurobiologically grounded framework that synergizes Personalized PageRank with retrieval-augmented generation, facilitating complex multi-hop reasoning through the integration of dense vector retrieval and sparse knowledge graph structures. \textbf{A-MEM}~\citep{xu2025mem} implements an agentic memory system inspired by the Zettelkasten method, enabling the dynamic construction and autonomous evolution of interconnected memory notes to refine knowledge representations over time. \textbf{Mem0}~\citep{chhikara2025mem0} provides a scalable memory architecture that dynamically extracts and consolidates conversational history into salient facts, supporting explicit operations to maintain long-term consistency in agentic interactions.

\subsection{Implementation Details}

We standardize the backbone models across all methods to ensure a fair comparison. We primarily employ Qwen3-Next-80B-A3B-Instruct~\cite{yang2025qwen3technicalreport} for both information extraction and downstream question answering tasks.To further validate the model-agnostic robustness and efficiency of the SEEM framework, we additionally incorporate Pangu-Embedded-7B~\cite{chen2025panguembeddedefficientdualsystem}, a compact model with 7 billion parameters, as a secondary backbone. This allows us to observe whether the episodic memory benefits scale down effectively to smaller-parameter models. Furthermore, cross-model validation using GPT-OSS-120B~\cite{agarwal2025gpt} is provided in Appendix~\ref{subsec:cross_model_validation}.
Regarding retrieval configurations, we align the hyperparameters based on the granularity of the retrieved units. For standard RAG baselines that rely on dense retrieval, we set the retrieval count $k$ to 5, fetching the top-5 original interaction messages. Similarly, for the memory-augmented baselines Mem0 and A-MEM, we retrieve the top-10 processed memory chunks. For HippoRAG 2, which operates on a session-based retrieval logic, we utilize the top-5 retrieved chunks to construct the context for final response generation. In our proposed SEEM framework, we configure the system to retrieve the top-5 relevant text chunks alongside their associated episodic memories to construct the reasoning context. To balance narrative continuity with information density, we employ a selective RPE strategy. Specifically, the total size of the final expanded evidence set $\mathcal{P}_{final}$ is restricted to at most twice the initial retrieval budget.

\section{Results}
\subsection{Main Results}
\label{subsec:main_results}

Table~\ref{tab:main_results} summarizes the performance of SEEM and several baseline methods on the LoCoMo and LongMemEval benchmarks, while Table~\ref{tab:category_breakdown} provides a detailed breakdown across different question categories. Overall, experimental results indicate that SEEM yields the highest scores across most evaluation metrics, reflecting its capacity for managing long-term agentic memory.

\paragraph{Comparison with Dense Retrieval.}
As shown in the first group of Table~\ref{tab:main_results}, while advanced dense retrieval models such as NV-Embed-v2 exhibit competitive performance in fetching local information, they remain limited by the absence of a structured memory state. SEEM exceeds the performance of NV-Embed-v2 by 3.2\% in F1 score and 3.3\% in LLM-as-a-Judge ($J$) score on LoCoMo. This performance gap suggests that pure vector-based retrieval, although efficient, may not fully capture the intricate relational and temporal dependencies of long-term interactions. By integrating structured EEFs and relational quadruples, SEEM provides a context that is more logically grounded compared to simple embedding-based matching.

\paragraph{Comparison with Memory-based Frameworks.}
SEEM consistently outperforms the evaluated memory-based systems across both benchmarks. On LoCoMo, SEEM achieves an F1 score of 61.1 and a $J$ score of 78.0, exceeding HippoRAG 2, by 2.8\% and 1.5\% respectively. It is observed that older memory frameworks yield lower performance scores, likely due to their reliance on flatter storage structures when processing extremely long interaction streams. In contrast, our hierarchical architecture facilitates an organized representation of complex event sequences. This trend is evident on LongMemEval, where SEEM achieves 65.0\% accuracy, representing a 4.4\% absolute improvement over HippoRAG 2.

\paragraph{Performance by Question Category.}
The categorical breakdown in Table~\ref{tab:category_breakdown} provides further insights into the framework's strengths. SEEM exhibits superior performance in four out of five categories, with notable gains in \textit{single-hop} and \textit{temporal} reasoning. The advantage in temporal queries suggests that the event-centric indexing within the episodic layer effectively maintains chronological narrative flow. Furthermore, SEEM achieves a high score in the \textit{adversarial} category, indicating that its provenance-based grounding helps distinguish factual evidence from distractors. Conversely, SEEM shows lower performance in the \textit{open-domain} category compared to HippoRAG 2. This suggests that for queries lacking specific narrative anchors, a purely graph-based retrieval approach without episodic expansion may be more efficient.

\paragraph{Semantic vs. Lexical Performance.}
A key observation is that the performance gains of SEEM are particularly evident in the LLM-as-a-Judge ($J$) and LongMemEval accuracy (Acc.) metrics. These metrics prioritize semantic alignment and factual correctness over surface-level word overlap (measured by BLEU-1). The scores in these categories indicate that SEEM does not merely retrieve relevant text but also reconstructs the underlying narrative logic. This synthesis is primarily driven by the RPE mechanism, which ensures that retrieved fragments are expanded into complete event contexts to support accurate reasoning.

\paragraph{Efficacy on Parameter-Constrained Backbones.} The results in Table~\ref{tab:pangu_results} reveal a compelling trend: SEEM effectively narrows the performance gap between small-scale models and their larger counterparts. Even when powered by Pangu-Embedded-7B, SEEM consistently outperforms all baseline frameworks, including those utilizing more complex retrieval-augmented strategies. Specifically, the significant lead in the reasoning assessment ($J$) on LoCoMo suggests that SEEM's episodic memory provides higher information density and narrative coherence, which is crucial for smaller LLMs that often struggle with "distraction" in long-context windows. Furthermore, the robust accuracy on LongMemEval confirms that SEEM does not rely on massive parameter counts to maintain long-term context; instead, it provides a specialized memory scaffolding that enables a 7B model to achieve reasoning depth previously reserved for much larger architectures. This highlights SEEM as a model-agnostic, parameter-efficient solution for long-range agentic memory.
\subsection{Hyperparameter Sensitivity Analysis}
\label{subsec:hyperparameter_analysis}

\begin{figure}[t]
    \centering
    \includegraphics[width=1.0\columnwidth]{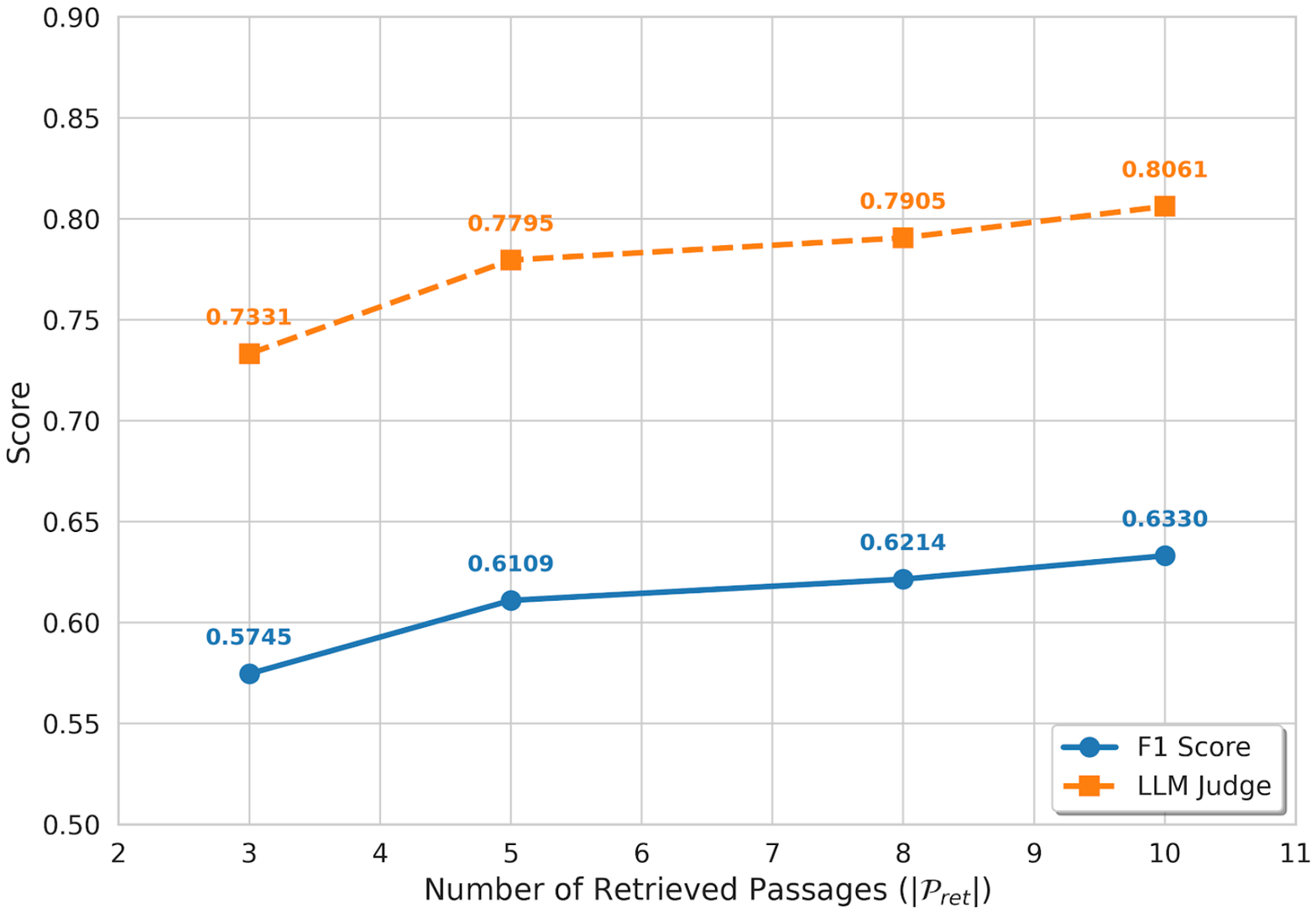} 
    \caption{Impact of the initial retrieval size ($|\mathcal{P}_{ret}|$).}
    \label{fig:ret_size_impact}
    \vspace{-4mm}
\end{figure}

We analyze the impact of the initial retrieval size $|\mathcal{P}_{ret}|$ on the reasoning performance of SEEM. This parameter controls the number of seed passages retrieved from the GML before any expansion occurs. Figure~\ref{fig:ret_size_impact} shows the performance trends for F1 and  $J$ as $|\mathcal{P}_{ret}|$ varies from 3 to 10.

We observe a consistent improvement in both metrics as the initial retrieval window expands. Specifically, increasing $|\mathcal{P}_{ret}|$ from 3 to 10 results in a 5.9\% gain in F1. Notably, SEEM does not exhibit the typical performance degradation often seen in traditional RAG systems when the context window grows. This positive correlation suggests that our framework can effectively leverage a broader range of initial evidence to refine its final answer without being overwhelmed by the additional potential noise in the retrieved passages.

\subsection{Ablation Study}

\begin{table}[t]
\centering
\small

\setlength{\tabcolsep}{8pt} % 保持适中的列间距
\resizebox{\columnwidth}{!}{% 适应单栏宽度
\begin{tabular}{l|ccc}
\toprule
\multirow{2}{*}{\textbf{Configuration}} & \multicolumn{3}{c}{\textbf{LoCoMo}} \\
 & BLEU-1 & F1 & $J$ \\
\midrule
\textbf{SEEM (Full Model)} & \textbf{56.1} & \textbf{61.1} & \textbf{78.0} \\
\midrule
\quad w/o Fact Provisioning ($\mathcal{K}_{top}$) & 55.2 & 60.4 & 77.7 \\
\quad w/o Relational Propagation & 54.5 & 59.6 & 76.3 \\
\quad w/o RPE & 55.1 & 60.2 & 77.1 \\
\quad w/o EEF ($\mathcal{E}_{ret}$) & 53.5 & 58.5 & 75.0 \\
\bottomrule
\end{tabular}%
}
\caption{Ablation study of key components in the SEEM framework on the LoCoMo benchmark.}
\vspace{-4mm}
\label{tab:ablation_study}
\end{table}

We conduct an ablation study to evaluate the individual contributions of the core components. We compare the full framework against four variants: (1) \textit{w/o Fact Provisioning}, which excludes the injection of relational quadruples; (2) \textit{w/o Relational Propagation}, which replaces the graph-based seed set expansion with direct lexical retrieval; (3) \textit{w/o RPE}, which disables the Reverse Provenance Expansion mechanism; and (4) \textit{w/o EEF}, which removes the structured episodic event frames.

\paragraph{Contribution of System Components.}
As shown in Table~\ref{tab:ablation_study}, the removal of any component leads to a measurable decrease across all evaluation metrics, confirming their synergy. The relational propagation mechanism serves as the foundation for identifying relevant historical entries; its absence results in a notable decline in the LLM Judge score, as the system struggles to navigate the global graph topology to locate non-contiguous passages. The RPE mechanism plays a key role in enriching the retrieved context; its absence leads to fragmented evidence, which negatively impacts reasoning quality. The omission of fact provisioning primarily impacts the factual grounding of responses, as the LLM lacks the explicit logical constraints provided by the graph-based quadruples. Finally, the EEFs provide the necessary structure for event-centric synthesis. The omission of EEFs requires the model to rely on unstructured text, which may lead to a decrease in the coherence of the generated responses.

\paragraph{Architectural Robustness.}
Experimental results indicate that SEEM maintains a consistent performance threshold even under ablated configurations. Observationally, the core hierarchical architecture yields reasoning scores that exceed those of established baselines, even when specific modules are deactivated. This suggests that the fundamental separation of episodic and relational information provides a structurally effective foundation for managing long-term context. These findings imply that the performance gains of SEEM are derived not only from auxiliary components but also from the underlying organization of its memory layers.

The results demonstrate that while the hierarchical architecture ensures a strong performance baseline, the integration of EEFs, RPE, fact provisioning, and relational propagation is essential to achieve optimal reasoning accuracy.

\subsection{Case Study}
\label{sec:case_study}

To qualitatively evaluate SEEM, we compare it against the gold standard and HippoRAG 2 on the LoCoMo (see Table~\ref{tab:case_study}). Our analysis focuses on three critical dimensions of agentic memory.

\paragraph{Multi-attribute Grounding.} 
Unlike raw text snippets, the EEF explicitly decomposes each interaction into granular roles such as \textit{Reason} and \textit{Method}. This structural decomposition allows the agent to distinguish between the intent and the action, which facilitates deeper social and causal reasoning across extended interaction histories.

\paragraph{Narrative Synthesis.} 
The framework achieves narrative synthesis through the \textit{Associative Fusion} of conversational turns. By merging an inquiry and its corresponding response into a single cohesive unit, the system effectively preserves the logical flow of the interaction. This consolidation approach also significantly reduces retrieval redundancy by avoiding the storage of fragmented conversational turns.

\paragraph{Temporal Resolution.} 
The frame exhibits sophisticated temporal grounding by processing reference dates alongside relative durations. For instance, by analyzing a reference date of January 23, 2022 in conjunction with a duration of three years, the system implicitly resolves the event's origin to January 2019. Such precise resolution ensures chronological consistency and factual accuracy within the EML.

In summary, SEEM ensures more grounded and logically consistent responses by transforming disparate interactions into a structured, coherent agentic memory.

\section{Conclusion}
\label{sec:conclusion}
We proposed SEEM, a hierarchical framework addressing scattered retrieval in long-term interactions. By integrating episodic event frames with an associative fusion mechanism, the system synthesizes coherent narratives from fragmented observations, outperforming traditional RAG and graph-based baselines. Our method effectively maintains global context and provides a scalable approach for enhancing the long-term reasoning capabilities of LLM-based agents in complex environments.

% Experimental results on the LoCoMo and LongMemEval benchmarks demonstrate that SEEM achieves state-of-the-art performance in multi-hop and temporal reasoning tasks compared to traditional RAG and graph-based baselines.
\section*{Limitations}

Despite its effectiveness, the framework faces limitations regarding computational efficiency, as the heavy reliance on LLMs for extracting frames and performing associative fusion increases latency and token costs compared to standard vector retrieval. Additionally, the system is susceptible to error propagation, where inaccuracies in the initial LLM-based extraction or fusion phases can permanently corrupt the structured memory store. Finally, the reliance on predefined semantic slots for event frames may limit the ability to capture abstract information that does not fit neatly into standard cognitive frame definitions.

\section*{Ethical Considerations}

The development of SEEM introduces considerations regarding the management and persistence of long-term interaction data. Unlike standard retrieval augmented generation which primarily accesses external corpora , SEEM transforms interaction streams into persistent episodic event frames and relational quadruples. While our experiments are conducted on publicly available benchmarks , real-world deployment of such a memory framework involves the retention of user information over extended periods. It is essential that future applications implement data anonymization protocols and provide users with explicit control over their stored interaction histories, including the right to modify or delete specific memory frames.

The framework is also subject to algorithmic bias and safety. Since SEEM relies on large language models for both episodic frame extraction and final response generation , it may inherit or amplify social biases present in these underlying models. The structured nature of event frames could potentially solidify these biases within the agent’s long-term memory, leading to biased reasoning in subsequent interactions. We recommend that developers implement content filtering and auditing mechanisms during the memory consolidation phase to mitigate these risks.
% Bibliography entries for the entire Anthology, followed by custom entries
%\bibliography{anthology,custom}
% Custom bibliography entries only
\bibliography{custom}

\clearpage
\appendix

\begin{table*}[t]
\centering
\small

\begin{tabular}{p{0.25\textwidth} p{0.20\textwidth} p{0.20\textwidth} p{0.20\textwidth}}
\toprule
\textbf{Query} & \textbf{Gold Answer} & \textbf{HippoRAG 2} & \textbf{SEEM (Ours)} \\
\midrule

% --- Case 1 ---
\textbf{Q1:} What book did Melanie read from Caroline's suggestion? (Multi-hop) & 
"Becoming Nicole" & 
The book's title is not specified. & 
"Becoming Nicole" by Amy Ellis Nutt \\
\midrule

% --- Case 2 ---
\textbf{Q2:} How did John describe his kids' reaction at the military memorial? (Single-hop)& 
Awestruck and humbled. & 
John said the experience made an impact on his kids, but did not describe their specific reaction. & 
They were awestruck and humbled. \\
\midrule

% --- Case 3 ---
\textbf{Q3:} What day did Tim get into his study abroad program? (Temporal) & 
January 5, 2024 & 
January 7, 2024 & 
January 5, 2024 \\

\bottomrule
\end{tabular}
\caption{Case study comparison between the gold answer and different memory frameworks.}

\label{tab:case_study}
\end{table*}

\section{Supplemental Experimental Results}\label{sec:supplemental}

\subsection{Cross-Model Generalization and Architectural Robustness}
\label{subsec:cross_model_validation}

To evaluate whether the performance gains of SEEM are model-dependent or derive from its underlying architecture, we conduct supplemental experiments on the LoCoMo benchmark by replacing the primary Qwen3-Next-80B-A3B-Instruct backbone with \textbf{GPT-OSS-120B}. This cross-model validation serves as a controlled comparison, ensuring that the observed improvements are attributable to our hierarchical memory mechanisms rather than the inherent capabilities of a specific LLM.

\begin{table*}[t]
\centering
\small
\setlength{\tabcolsep}{10pt}

\resizebox{0.7\textwidth}{!}{%
\begin{tabular}{lccc}
\toprule
\multirow{2}{*}{\textbf{Method}} & \multicolumn{3}{c}{\textbf{LoCoMo}} \\
\cmidrule(lr){2-4}
 & BLEU-1 & F1 & $J$ \\
\midrule

% --- 第一类：Dense Retrieval ---
\rowcolor{gray!5}
\multicolumn{4}{l}{\textit{Dense Retrieval}} \\
KaLM-Embedding-V2.5~\citep{zhao2025kalmembeddingv2superiortrainingtechniques} & 38.7 & 42.8 & 63.2 \\ 
NV-Embed-v2~\citep{lee2025nvembed} & 44.1 & 49.2 & 75.5 \\
\midrule

% --- 第二类：Memory-based Frameworks ---
\rowcolor{gray!5}
\multicolumn{4}{l}{\textit{Memory-based Frameworks}} \\
A-MEM~\citep{xu2025mem} & 42.4 & 47.3 & 63.0 \\
HippoRAG 2~\citep{gutierrez2025from} & 44.6 & 50.2 & 73.6 \\

\rowcolor{gray!15} 
\textbf{SEEM (Ours)} & \textbf{50.7} & \textbf{55.7} & \textbf{77.1} \\
\midrule
\multicolumn{4}{l}{\textit{Backbone LLM: GPT-OSS-120B}} \\
\bottomrule
\end{tabular}%
}
\caption{Performance comparison on LoCoMo based on GPT-OSS-120B. The best results are highlighted in \textbf{bold}.}
\label{tab:main_results_oss}
\end{table*}

\paragraph{Analysis of Results.} 
As demonstrated in Table~\ref{tab:main_results_oss}, SEEM maintains its performance leadership when integrated with the GPT-OSS-120B backbone, mirroring the trends observed with the Qwen3-Next-80B-A3B-Instruct model. The consistent performance gains across these distinct large language models reinforce the conclusion that the advantages of our hierarchical episodic architecture are model-agnostic. By decoupling the memory organization mechanism from the specific underlying LLM, SEEM demonstrates robust generalization capabilities in narrative consistency and retrieval precision. These results confirm that the framework serves as a versatile enhancement for various long-context reasoning agents regardless of their specific architectural implementations.

\subsection{Granular Category-wise Evaluation}
\label{subsec:category_analysis}

To further investigate the performance characteristics of SEEM across diverse reasoning challenges, we present a granular analysis of the results on the \textbf{LoCoMo} and \textbf{LongMemEval} benchmarks, categorized by specific task dimensions.

\paragraph{Analysis of LoCoMo Categories.} 
As illustrated in Table~\ref{tab:category_correctness}, SEEM achieves superior performance across four out of five reasoning categories. The framework demonstrates significant advantages in Temporal and Multi-hop reasoning, outperforming competitive baselines by a notable margin. These results suggest that the structured EEFs effectively capture chronological dependencies that are often overlooked by dense retrieval or static graph-based approaches. While HippoRAG 2 maintains competitive performance in Open-domain queries due to its focus on static entity indexing, SEEM prioritizes the reconstruction of complex narrative chains. This architectural focus is further evidenced by SEEM's higher resilience to adversarial distractors, indicating lower vulnerability to hallucinations compared to traditional retrieval-based systems.

\begin{table*}[t]
\centering
\small
\setlength{\tabcolsep}{6pt}
\resizebox{\textwidth}{!}{%
\begin{tabular}{lcccccccccc}
\toprule
\multirow{3}{*}{\textbf{Method}} & \multicolumn{2}{c}{\textbf{Multi-hop}} & \multicolumn{2}{c}{\textbf{Temporal}} & \multicolumn{2}{c}{\textbf{Open-domain}} & \multicolumn{2}{c}{\textbf{Single-hop}} & \multicolumn{2}{c}{\textbf{Adversarial}} \\
 & \multicolumn{2}{c}{(Count: 282)} & \multicolumn{2}{c}{(Count: 321)} & \multicolumn{2}{c}{(Count: 96)} & \multicolumn{2}{c}{(Count: 841)} & \multicolumn{2}{c}{(Count: 446)} \\
\cmidrule(lr){2-3} \cmidrule(lr){4-5} \cmidrule(lr){6-7} \cmidrule(lr){8-9} \cmidrule(lr){10-11}
 & Correct & Acc. & Correct & Acc. & Correct & Acc. & Correct & Acc. & Correct & Acc. \\
\midrule
A-MEM & 154 & 54.61\% & 90 & 28.04\% & 45 & 46.88\% & 496 & 58.98\% & 430 & 96.41\% \\
HippoRAG 2 & 173 & 61.35\% & 203 & 63.24\% & \textbf{59} & \textbf{61.46\%} & 659 & 78.36\% & 416 & 93.27\% \\
NV-Embed-v2 & 148 & 52.48\% & 205 & 63.86\% & 56 & 58.33\% & 647 & 76.93\% & 417 & 93.50\% \\
\rowcolor{gray!15}
\textbf{SEEM (Ours)} & \textbf{177} & \textbf{62.77\%} & \textbf{219} & \textbf{68.22\%} & 52 & 54.17\% & \textbf{668} & \textbf{79.43\%} & \textbf{432} & \textbf{96.86\%} \\
\bottomrule
\end{tabular}%
}
\caption{Category-specific performance on the LoCoMo dataset. Sample counts for each reasoning category are provided in parentheses. The best results are highlighted in \textbf{bold}.}
\label{tab:category_correctness}
\end{table*}

% \begin{figure}[htbp]
%     \centering
%     \includegraphics[width=0.85\linewidth]{figures/LoCoMo_Category.png}
%     \caption{Performance comparison across five reasoning categories in the LoCoMo benchmark.}
%     \label{fig:radar_locomo}
% \end{figure}

\paragraph{Analysis of LongMemEval Categories.} 
The evaluation encompasses six distinct reasoning categories: Speaker-Specific (S-S) tasks focused on the user, assistant, or preferences; Multi-Session (Multi-S) interaction; Temporal reasoning; and Knowledge Update (K-Update). This comprehensive assessment further reinforces the efficacy of the SEEM architecture. As shown in Table~\ref{tab:longmemeval_horizontal}, SEEM achieves the highest average accuracy, driven primarily by its strong performance in the Knowledge Update and Temporal reasoning categories. The framework’s capacity to resolve user-specific information highlights its effectiveness in grounding queries to the appropriate episodic context. While certain baselines demonstrate specialized strengths in preference-based retrieval, SEEM provides a more balanced performance profile. This equilibrium is achieved by bridging high-level semantic abstractions with the granular requirements of long-term interaction history, ensuring consistent reasoning across diverse and evolving query types.

\begin{table*}[t]
\centering
\small
\setlength{\tabcolsep}{5pt}
\resizebox{\textwidth}{!}{%
\begin{tabular}{lccccccc}
\toprule
\textbf{Method} & \makecell{\textbf{S-S (User)}\\(Count: 70)} & \makecell{\textbf{S-S (Asst.)}\\(Count: 56)} & \makecell{\textbf{S-S (Pref.)}\\(Count: 30)} & \makecell{\textbf{Multi-S}\\(Count: 133)} & \makecell{\textbf{Temporal}\\(Count: 133)} & \makecell{\textbf{K-Update}\\(Count: 78)} & \textbf{Mean} \\
\midrule
HippoRAG 2 & 82.86 & \textbf{94.64} & 20.00 & \textbf{58.65} & 48.12 & 56.41 & 60.11 \\
NV-Embed-v2 & 80.00 & \textbf{94.64} & \textbf{33.33} & 48.12 & 43.61 & 65.38 & 60.85 \\
\rowcolor{gray!15}
\textbf{SEEM (Ours)} & \textbf{91.43} & \textbf{94.64} & 30.00 & 54.89 & \textbf{53.38} & \textbf{70.51} & \textbf{65.81} \\
\bottomrule
\end{tabular}%
}
\caption{Detailed performance comparison on the LongMemEval benchmark. Accuracy (\%) is reported across six reasoning categories, with sample counts for each category provided in parentheses. The best results are highlighted in \textbf{bold}.}
\label{tab:longmemeval_horizontal}
\end{table*}

% \begin{figure}[htbp]
%     \centering
%     \includegraphics[width=0.85\linewidth]{figures/LongMemEval_Category.png}
%     \caption{Performance comparison of SEEM against HippoRAG 2 and NV-Embed-V2 baselines across the six evaluation categories of the LongMemEval benchmark.}
%     \label{fig:longmemeval_radar}
% \end{figure}

\subsection{Evaluation of Incremental Memory Construction}
To assess the practical applicability of SEEM in streaming interaction scenarios, we conduct an evaluation under an incremental construction setting. In this configuration, the complete sequence of interaction passages is partitioned into four chronological segments, which are processed by the memory system sequentially rather than in a single batch.

The results, summarized in Table~\ref{tab:incremental_test}, demonstrate that SEEM maintains highly stable performance across all evaluation metrics. The marginal discrepancy between the batch and incremental modes suggests that the associative fusion mechanism effectively preserves narrative coherence and structural integrity, even when information is presented in fragments. This minimal performance trade-off confirms the framework's robustness for real-world deployment, where memory must evolve continuously in response to sequential updates without significant loss in reasoning integrity.
\begin{table}[htbp]
\centering
\small % Slightly smaller font to fit better

\resizebox{\columnwidth}{!}{% Use resizebox to force the table into the column width
\begin{tabular}{lccc}
\toprule
\textbf{Method} & \textbf{BLEU-1} & \textbf{F1} & \textbf{$J$} \\ \midrule
SEEM (Batch) & \textbf{56.1} & \textbf{61.1} & \textbf{78.0} \\
SEEM (Incremental) & 55.6 & 60.6 & 77.6 \\ \bottomrule
\end{tabular}
}
\caption{Comparison between Batch and Incremental Memory Construction in SEEM.}
\label{tab:incremental_test}
\end{table}

\section{Analysis}\label{appendix:analysis}
\subsection{Structural Analysis of the Graph Memory Layer}
\label{subsec:graph_layer_analysis}

The GML provides the static factual foundation of the SEEM framework, complementing the dynamic nature of the EML. As summarized in Table~\ref{tab:graph_layer_transposed}, the structural statistics across various narrative partitions reflect a high density of relational knowledge and entity connectivity.

The internal composition of the graph highlights two critical capabilities of the system. The prevalence of temporal anchors indicates that a vast majority of the extracted facts are grounded in specific temporal contexts, which is essential for resolving chronological dependencies in long-term reasoning. This structural density ensures that the GML can serve as a reliable foundation for relational propagation, providing the necessary factual context for hybrid retrieval.
\begin{table*}[t]
\centering
\small

\setlength{\tabcolsep}{4.8pt} % 略微调整列间距以完美适配页面
\begin{tabular}{lccccccccccc}
\toprule
\textbf{Metric} & $h_1$ & $h_2$ & $h_3$ & $h_4$ & $h_5$ & $h_6$ & $h_7$ & $h_8$ & $h_9$ & $h_{10}$ & \textbf{Average} \\
\midrule
Entities        & 1,242 & 902   & 1,845 & 1,486 & 1,820 & 1,692 & 1,745 & 1,665 & 1,286 & 1,575 & \textbf{1,525.8} \\
Facts   & 1,749 & 1,320 & 2,534 & 2,194 & 2,673 & 2,699 & 2,557 & 2,395 & 1,868 & 2,348 & \textbf{2,233.7} \\
Temporal Anchors& 1,557 & 1,213 & 2,294 & 1,948 & 2,363 & 2,385 & 2,258 & 2,070 & 1,694 & 2,056 & \textbf{1,983.8} \\
Synonymy Edges  & 11,732 & 5,439   & 19,963 & 14,178 & 16,433 & 14,344 & 15,904 & 15,402 & 10,670 & 12,459 & \textbf{13,652.4} \\
\bottomrule
\end{tabular}
\caption{Structural statistics of the GML across 10 Narrative Partitions ($h_1 \text{--} h_{10}$) in the LoCoMo dataset. The metrics quantify the internal density of the GML, representing the static knowledge foundation of the SEEM framework.}
\label{tab:graph_layer_transposed}
\end{table*}

\subsection{Qualitative Analysis of Episodic Event Frames}
\label{subsec:eef_qualitative}

Figure~\ref{fig:example_eef} provides a representative instance of a consolidated EEF, illustrating the framework's capacity for high-fidelity narrative synthesis. Several key advantages of the SEEM architecture are evident in this structured representation:

\paragraph{Multi-attribute Grounding.} 
Unlike raw text snippets, the EEF explicitly decomposes the interaction into fine-grained roles such as Reason and Method. This decomposition allows the agent to distinguish between intent and action, which facilitates deeper social and causal reasoning across extended interaction histories.

\paragraph{Narrative Synthesis.} 
The SEEM framework achieves narrative synthesis through the associative fusion of interaction pairs. By merging a conversational inquiry and its corresponding response into a single, cohesive episodic unit, the system preserves the logical continuity of the dialogue. This consolidation mechanism effectively captures the functional relationship between speaker turns while significantly reducing retrieval redundancy in the memory store.

\paragraph{Temporal Resolution.} 
The EEF exhibits sophisticated temporal grounding by processing the reference date alongside the relative duration. For instance, the system implicitly resolves an event's origin to ``January 2019'' by analyzing the reference date in conjunction with a three-year duration. This precise resolution ensures chronological consistency and factual integrity within the EML.

By transforming ambiguous pronouns into structured attributes while maintaining strict textual grounding via provenance pointers, the EEF provides a high-density semantic anchor. This structured representation ensures that retrieved context is not only chronologically accurate but also logically complete for downstream reasoning.
% --- Consolidated Memory Example (Fixed Layout) ---
\begin{figure*}[t] % 使用星号 * 实现双栏跨栏排版
\begin{tcolorbox}[
    colback=blue!3!white, 
    colframe=blue!75!black, 
    title=\textbf{Example: Consolidated Episodic Memory Frame},
    fonttitle=\small\bfseries, 
    fontupper=\small, 
    arc=2pt, 
    outer arc=2pt
    % 已移除 breakable 以确保在浮动体内部显示完整且无横向分割线
]
\textbf{Summary:} On January 23, 2022, at 2:01 pm, Joanna asked Nate how long he had had ``them,'' prompting Nate to respond that he had owned them for three years—since approximately January 2019—and that they brought him significant joy.

\vspace{0.8em}
\textbf{Events (Structured EEF):}
\begin{itemize}[leftmargin=0pt, label={}, itemsep=1.2em] % 移除外部列表缩进
    \item \textbf{\underline{Event 1:}} \vspace{0.3em}
          \begin{description}[style=multiline, leftmargin=6em, font=\itshape, nosep]
            \item[Participants] Joanna
            \item[Action] Joanna asked how long Nate had had ``them''
            \item[Time] 2:01 pm on 23 January, 2022
            \item[Reason] Expressing affectionate curiosity about an object Nate possesses
            \item[Method] Through verbal inquiry
          \end{description}
          
    \item \textbf{\underline{Event 2:}} \vspace{0.3em}
          \begin{description}[style=multiline, leftmargin=6em, font=\itshape, nosep]
            \item[Participants] Nate
            \item[Action] Nate stated he had owned ``them'' for three years, and that they brought him tons of joy
            \item[Time] From approx. January 2019 to January 23, 2022
            \item[Reason] Responding to Joanna's question about the duration of ownership
            \item[Method] Through verbal response
          \end{description}
\end{itemize}
\end{tcolorbox}
\caption{An illustrative example of a consolidated Episodic Event Frame (EEF) in the SEEM framework. This structured representation demonstrates how the associative fusion mechanism synthesizes multi-turn interactions into coherent, attribute-rich episodic units.}
\label{fig:example_eef}
\end{figure*}
\subsection{Analysis of Associative Fusion}
\label{subsec:fusion_analysis}
We evaluate the structural impact of the associative fusion mechanism by analyzing the distribution of consolidated frames relative to the original interaction turns. As demonstrated in Table~\ref{tab:memory_fusion_stats}, SEEM reduces the total number of memory units by synthesizing fragmented turns into unified episodic frames. This consolidation mitigates semantic redundancy and improves retrieval density by grouping chronologically and logically linked interactions. The presence of multi-turn fusions indicates that the framework can bridge narrative sequences, transforming discrete conversational segments into more compact semantic representations. This structural efficiency ensures a logically continuous memory state, which is essential for maintaining context during long-horizon agentic reasoning.

\begin{table}[H]
\centering
\small

\begin{tabular}{cc}
\toprule
\makecell{\textbf{Passages per} \\ \textbf{Memory}} & \makecell{\textbf{Number of} \\ \textbf{Memory Frames}} \\
\midrule
1  & 371 \\
2  & 79 \\
3  & 20 \\
4  & 3 \\
5  & 4 \\
8  & 1 \\
\midrule
\textbf{Total Memory Frames} & \textbf{478} \\
\textbf{Total Passages} & \textbf{629} \\
\textbf{Consolidation Ratio} & \textbf{1.32:1} \\
\bottomrule
\end{tabular}
\caption{Distribution of consolidated episodic memory frames across constituent interaction passages in a LoCoMo narrative partition.}
\label{tab:memory_fusion_stats}
\end{table}

\subsection{Redundancy Analysis of Dual-Layer Retrieval}
\label{appendix:redundancy_analysis}

To verify the necessity of the dual-layer architecture, we analyze the global distribution of semantic redundancy between the GML and the EML. For each query in the LoCoMo dataset, we retrieve the corresponding structural quadruples from the GML and EEFs from the EML. We apply an LLM-based filter to the GML outputs to ensure precision, resulting in 1,282 valid retrieval pairs from the original 1,986 queries. The aggregate semantic overlap is quantified by computing the cosine similarity between their respective embeddings, with the overall distribution detailed in Table~\ref{tab:similarity_distribution}.

\begin{table}[H]
\centering
\small
\begin{tabular}{lrr}
\toprule
\textbf{Similarity Range} & \textbf{Count} & \textbf{Prop. (\%)} \\
\midrule
$[0.25, 0.30)$ & 1   & 0.08  \\
$[0.30, 0.35)$ & 12  & 0.94  \\
$[0.35, 0.40)$ & 106 & 8.27  \\
$[0.40, 0.45)$ & 398 & 31.05 \\
$[0.45, 0.50)$ & 497 & 38.77 \\
$[0.50, 0.55)$ & 224 & 17.47 \\
$[0.55, 0.60)$ & 40  & 3.12  \\
$[0.60, 0.65)$ & 4   & 0.31  \\
\midrule
\multicolumn{2}{c}{\textbf{Total Valid Pairs}} & \textbf{1282} \\
\multicolumn{2}{c}{\textbf{Mean Similarity}}  & \textbf{0.46} \\
\bottomrule
\end{tabular}
\caption{Distribution of cosine similarity between retrieved quadruples (GML) and EEFs (EML) on the LoCoMo dataset.}
\label{tab:similarity_distribution}
\end{table}

The mean similarity of 0.46 suggests that the GML and EML capture complementary semantic dimensions. This divergence confirms that the structural extraction and narrative synthesis capture distinct information even when grounded in the same interaction context, justifying the use of a dual-layer architecture.

\section{Prompt Templates and Agent Instructions}
\label{appendix:prompts}

In this section, we provide the detailed prompt templates used in the SEEM framework. These prompts are designed to implement the formal functions defined in Section~\ref{sec:methodology}, specifically the extraction function $\mathcal{F}_{ext}$, the consolidation function $\mathcal{F}_{fuse}$, and the final generation function $G$.

% 请确保导言区包含：\usepackage{tcolorbox, enumitem}

\begin{figure*}[t] % 星号 * 确保跨越双栏
\begin{tcolorbox}[
    colback=gray!5, 
    colframe=gray!75, 
    title=\textbf{Prompt 1: Episodic Event Frame Extraction ($\mathcal{F}_{\text{ext}}$)},
    fonttitle=\small\bfseries, 
    fontupper=\small, 
    arc=2pt,
    outer arc=2pt

]
You are an expert at extracting episodic memories from conversation turns. Your task is to analyze a single conversation turn (which may contain time information, speaker information, and text content), identify distinct events mentioned in the turn, and extract structured event details for each event.\\[0.5em]

The input format is a single conversation turn that may include:
\begin{itemize}[leftmargin=1.5em, noitemsep, topsep=2pt]
    \item Time information
    \item Image descriptions in the format: \texttt{[Image: <description>]}
    \item The actual text content of the turn
\end{itemize}

For each event, extract the following Structured Event Attributes (use \texttt{null} if not reliably determined):
\begin{itemize}[leftmargin=1.5em, noitemsep, topsep=2pt]
    \item \textbf{Participants:} List of actors. Replace pronouns with specific names; include full names/roles.
    \item \textbf{Action:} List of substantive actions. \textit{CRITICAL}: Each action MUST include the subject/actor. Format: ``Subject verb object''.
    \item \textbf{Time:} Time, date, or duration. For continuous events, explicitly specify the range. Specify the event's actual time, not just the conversation time.
    \item \textbf{Location/Reason/Method:} The venue, purpose, or means if explicitly stated.
\end{itemize}

\textbf{Guidelines:}
1) \textbf{Event Definition:} Define an ``event'' as an occurrence with a clear subject, action, and temporal context. 2) \textbf{Coreference Resolution:} Resolve pronouns to specific entities. 3) \textbf{No Redundancy:} Do not extract the act of ``speaking'' as a separate event. 4) \textbf{Output strict JSON.} No markdown formatting.\\[0.5em]

\textbf{Output format (strict JSON):}
\begin{tcolorbox}[colback=white, colframe=gray!30, arc=1pt, boxrule=0.5pt, left=2pt, right=2pt, top=2pt, bottom=2pt]
\begin{flushleft}
\small\ttfamily
\{ \\
\hspace*{1em}"summary": "1-3 concise sentences...", \\
\hspace*{1em}"events": [\{ "participants": [], "action": [], "time": "", ... \}] \\
\}
\end{flushleft}
\end{tcolorbox}
\end{tcolorbox}
\caption{The structured prompt for Episodic Event Frame Extraction ($\mathcal{F}_{ext}$). This initial stage of the SEEM pipeline converts unstructured interaction logs into discrete, attribute-rich event units, providing the grounded anchors necessary for long-term temporal and multi-hop reasoning.}
\label{fig:prompt_extraction}
\end{figure*}

% --- Prompt 2: Consolidation ---
\begin{figure*}[t] 
\begin{tcolorbox}[
    colback=gray!5, 
    colframe=gray!75, 
    title=\textbf{Prompt 2: Associative Consolidation and Fusion ($\mathcal{F}_{\text{fuse}}$)},
    fonttitle=\small\bfseries, 
    fontupper=\small, 
    arc=2pt,
    outer arc=2pt,
]
You are an expert at integrating episodic memories. Your task is to combine two episodic memories into a single, comprehensive memory that preserves all important information from both memories.\\

\textbf{IMPORTANT: Event Integration Strategy}
\begin{itemize}[leftmargin=1.5em, noitemsep, topsep=2pt]
    \item \textbf{LESS:} When events from both memories describe the same occurrence (merge).
    \item \textbf{EQUAL:} When all events are distinct and unrelated.
    \item \textbf{GREATER:} When integration reveals new event relationships.
\end{itemize}

\textbf{Guidelines:}
\begin{enumerate}[leftmargin=1.5em, noitemsep, topsep=1pt]
    \item \textbf{Conservative Merge:} Only merge events that clearly describe the EXACT SAME occurrence. If events are part of a sequence (Plan $\rightarrow$ Execute), DO NOT merge them.
    \item \textbf{Entity Alignment:} Unify participant names.
    \item \textbf{Conflict Resolution:} Highest Priority: Evidence found in ``Original Passages''. Use these original sources as the primary reference to cross-verify all episodic attributes and mitigate error propagation.
    \item \textbf{Summary Synthesis:} Do NOT simply concatenate. Rewrite a single, cohesive narrative describing progression or causal relationships.
\end{enumerate}

\textbf{CRITICAL: Time Information Handling Rules}
\begin{enumerate}[leftmargin=1.5em, noitemsep, topsep=1pt]
    \item \textbf{Same Event, Different Time:} If one is more specific, prefer it; if complementary, combine into a range. 
    \item \textbf{Sequential Events:} Events at different times MUST be kept separate. 
    \item \textbf{Temporal Ordering:} Arrange events in chronological order.
\end{enumerate}

\textbf{Output Format:} Strict JSON ONLY.
\end{tcolorbox}
\caption{The structured prompt for associative consolidation and fusion, designed to synthesize fragmented interaction logs into coherent Episodic Event Frames (EEFs).}
\label{fig:prompt_fusion_full}
\end{figure*}
% Required packages: \usepackage{tcolorbox, enumitem}

\begin{figure*}[t] % Standard environment for full-width objects in double-column papers
\begin{tcolorbox}[
    colback=gray!5, 
    colframe=gray!75, 
    title=\textbf{Prompt 3: Memory-Augmented Question Answering ($G$)},
    fonttitle=\small\bfseries, 
    fontupper=\small, 
    arc=2pt, 
    outer arc=2pt
    % 'breakable' is removed to prevent internal segmentation lines
]
You are a reading-comprehension QA assistant operating in episodic-memory mode.\\

Each query provides:
\begin{itemize}[leftmargin=2em, noitemsep, topsep=2pt]
    \item[(A)] an ``Original Passages (Grounded Evidence)'' section (most trusted evidence);
    \item[(B)] an ``Episodic Memory Summary'' section (high-signal reference distilled from retrieved passages; use as a guide and as supplemental evidence when not contradicted);
    \item[(C)] optionally, a ``Relevant Facts'' section (high-signal reference quadruples; use to locate/verify key entities/relations and as supplemental evidence when not contradicted).
\end{itemize}

\textbf{Evidence Policy:}
\begin{enumerate}[leftmargin=2em, noitemsep, topsep=2pt]
    \item You MUST read (B) and (C) (if present). Treat them as high-signal hints to guide what to look for in (A).
    \item If (A) is incomplete, you MAY answer using explicit (B)/(C) as supplemental references ONLY if they do not contradict (A).
    \item If (B)/(C) conflicts with (A), trust (A) and ignore the conflicting parts of (B)/(C).
\end{enumerate}

\textbf{Output Format:}\\
Thought: <reasoning>\\
Answer: <answer only>\\
(Constraint: The answer must be concise, definitive, and devoid of additional elaborations)
\end{tcolorbox}
\caption{The Inference Prompt for SEEM's Memory-Augmented Question Answering. By providing the model with distilled episodic summaries and graph-based facts alongside raw evidence, the system effectively mitigates the ``scattered retrieval'' problem in long-context interactions.}
\label{fig:prompt_inference}
\end{figure*}

\end{document}